\pdfoutput=1 
\documentclass[10pt,conference,a4paper]{IEEEtran}
\usepackage[top=0.705in, left=0.514in, bottom=1.605in, right=0.514in]{geometry}
\usepackage{todonotes}

\usepackage[hyphens]{url}
\usepackage{hyperref}
\usepackage[hyphenbreaks]{breakurl}
\usepackage{doi}

\usepackage{cite}

\ifCLASSINFOpdf
\else
\fi

\usepackage{amsmath}
\PassOptionsToPackage{hyphens}{url}\usepackage{hyperref}%\usepackage{hyperref}

\begin{document}
\bstctlcite{IEEEexample:BSTcontrol}

\title{ResNet-like Architecture with Low Hardware Requirements}

\author{\IEEEauthorblockN{Elena Limonova}
\IEEEauthorblockA{FRC CSC RAS,\\
Smart Engines Service LLC,\\
Moscow, Russia \\
Email: limonova@smartengines.com}
\and
\IEEEauthorblockN{Daniil Alfonso}
\IEEEauthorblockA{JSC MCST,\\
Moscow, Russia \\
Email: alfonso\_d@mcst.ru}
\and
\IEEEauthorblockN{Dmitry Nikolaev}
\IEEEauthorblockA{Smart Engines Service LLC,\\
Institute for Information Transmission\\
Problems RAS, Moscow, Russia\\
Email: dimonstr@iitp.ru }
\and
\IEEEauthorblockN{Vladimir V. Arlazarov}
\IEEEauthorblockA{FRC CSC RAS,\\
Smart Engines Service LLC,\\
Moscow, Russia \\
Email: vva@smartengines.com}
}

\maketitle

\begin{abstract}
One of the most computationally intensive parts in modern recognition systems is an inference of deep neural networks that are used for image classification, segmentation, enhancement, and recognition.
The growing popularity of edge computing makes us look for ways to reduce its time for mobile and embedded devices.
One way to decrease the neural network inference time is to modify a neuron model to make it more efficient for computations on a specific device.
The example of such a model is a bipolar morphological neuron model.
The bipolar morphological neuron is based on the idea of replacing multiplication with addition and maximum operations.
This model has been demonstrated for simple image classification with LeNet-like architectures \cite{limonova2019}.
In the paper, we introduce a bipolar morphological ResNet (BM-ResNet) model obtained from a much more complex ResNet architecture by converting its layers to bipolar morphological ones.
We apply BM-ResNet to image classification on MNIST and CIFAR-10 datasets with only a moderate accuracy decrease from 99.3\% to 99.1\% and from 85.3\% to 85.1\%. We also estimate the computational complexity of the resulting model. We show that for the majority of ResNet layers, the considered model requires 2.1-2.9 times fewer logic gates for implementation and 15-30\% lower latency. 
\end{abstract}


% \fi
\IEEEpeerreviewmaketitle

\section{Introduction}

Machine vision is becoming very popular and finds many practical applications \cite{8999509, chang2019vision, zhang2019deeper, li2018, bezmaternykh2019unetbin, bokovoy}. The proliferation of complex technical systems leads to the emergence of very different requirements for the computational algorithms used. The concept of edge computing is gaining popularity, which means that calculations are performed as close as possible to the end-user. In this case, it is often impossible to use powerful hardware, and the methods must work quickly and accurately enough on a device with limited resources. Therefore, the task of increasing the computational efficiency of pattern recognition algorithms is becoming more and more critical.

A widely used approach to pattern recognition relies on neural network-based algorithms.
The choice of specific neural network architecture can be pretty tricky and depends on the problem, the desired accuracy, and suitable computational efficiency \cite{vizilter2019structure}.
Cutting edge neural network architectures significantly differ from each other but have one thing in common.
They all strongly rely on convolutional layers.
These layers perform the convolution of the input signal with one or more filters.
The convolution operation has several features that are very important for visual recognition.
The first feature is the fact that the result does not depend on the spatial position of the image object.
The relatively small size of the filter provides an analysis of a small space region of image (receptive field) and allows us to select elementary features, for example, corners.
These elementary features are, in turn, analyzed by subsequent layers.
The aspects of convolutional neural networks repeat the properties of receptive neurons.

However, the computational operation performed inside the neuron is not limited to a weighted sum of input signals, followed by non-linearity.
For example, retinal neurons are ON~/~OFF neurons; that is, ON-neurons only work in the presence of light, and OFF-neurons activate in the dark.
Although we can model a similar effect using activation functions, modern neural network architectures typically include only a positive data processing pathway via Rectified Linear Unit~(ReLU) activation.
However, in \cite{kim} Kim et al. show that processing negative information and using separate signal processing paths can improve recognition accuracy.
Thus, there is a reason to believe that the traditional model of the convolutional neural network can be improved, considering the biological mechanisms of perception. 

The new bipolar morphological neuron model \cite{limonova2019} has all these features of perception, including separate processing paths for positive and negative data, that emulate excitation and inhibition inside the neuron. There is one more important thing to mention: it does not use multiplication in the operation of convolution, and uses addition and maximum instead. Therefore, there is reason to believe that such a model can be significantly more computationally efficient. Although it utilizes quite difficult activation functions outside the convolution, we can implement them efficiently for cases when we can use approximations. However, the model has been demonstrated only in image classification with LeNet-like networks. Compact lightweight networks are handy for mobile and embedded recognition, but there are a lot more tasks which require deep models to reach state-of-the-art quality.

In this paper, we demonstrate a bipolar morphological network of ResNet architecture \cite{he2016deep, he2016identity}. ResNet is based on residual blocks that allow stacking to obtain a deep neural network. The resulting network can solve many recognition problems with high quality and is very scalable to keep the desired balance between inference speed and accuracy. We show bipolar morphological ResNet-22 accuracy in image classification on CIFAR-10 and MNIST datasets. Thus, we demonstrate for the first time that bipolar morphological neurons can be used in deep neural networks. We also analyze the computational complexity of the network and estimate the number of logic gates required for implementation.

The structure of the paper is as follows.
Section 2 gives related work on network structures and speedup methods.
In Section 3, we describe the bipolar morphological model, and its complexity in terms of computing operations, logic gates and clock cycle latency.
Section 4 shows our experimental results on performance for the ResNet-like model on MNIST and CIFAR-10 image classification problems.
Finally, in Section~5, we summarize our paper.

\section{Related work}

Researchers pay more and more attention to various approximations of neural network structures and the development of alternative computational models.
The main problem solved in this way is the acceleration of the network inference on specific devices, which can vary significantly in their architecture and requires different approaches to optimize efficiency.

For example, a universal method is to reduce the redundancy of neural network architecture.
Low-rank approximations of computations are very popular in order to reduce the computational complexity of convolutional layers.
They are based on various decompositions of convolutional filters (SVD variations, for example) \cite {denton, jaderberg, rigamonti}, including depth-wise separable convolution and its modifications that divide the channels of the input image into groups and process them separately~\cite {chollet, limonova}.
These approaches are still being improved. For example, in \cite {Scheidegger}, authors propose separable convolutional eigen-filters, which reduce computational complexity and demonstrate the complete absence of loss in recognition quality.
In \cite {wang}, authors speed up convolutional layers by using depth-wise separable convolution and show a way to find the optimal channel configuration in groups to prevent a quality decrease.
Modern methods also combine different approaches, for example, low-rank approximations and sparse filters, and reduce the number of parameters of modern architectures by about 30\%, leaving accuracy almost at the same level \cite{guo}.

Low precision integer approximations are also very popular.
For example, in \cite {Yao}, authors propose an end-to-end 8-bit integer model without internal conversions to floating-point data types.
As a result, this model can provide fast inference on mobile and embedded devices with relatively small accuracy losses.
The research on increasing the accuracy of the quantized network is also underway \cite {choukroun2019low, pietron2019methodologies, sun2019multi, ilin}.
Among such approximations, binary neural network models \cite{mocerino, Li} occupy a special place.
They can be very computationally efficient on special devices and use less memory.
However, they are still inferior to floating-point models in terms of recognition accuracy. Fast implementations are also of interest \cite{limonova2020special, 10.1145/3306202}.

Recently, the idea of multiplication-free neural networks is gaining popularity again.
The first model without multiplications was a morphological neural network, which was proposed back in 1996 by Ritter \cite{Ritter} and then complemented with dendrites \cite{Ritter2003_dendrites}. It did not find extensive use due to the insufficiently high accuracy for complex recognition tasks.
Recently, however, new models have been proposed that restrict the use of multiplications, for example, DeepShift \cite{elhoushi2019deepshift}, where Elhoushi et al. replace multiplications by an effective bit shift.
In \cite {chen2019addernet}, Chen et al. propose to use the $L_1$-norm in convolutional layers, while preserving the multiplications in the batch-normalization layer.
To train the network, the authors used sign gradient in backpropagation.
The results demonstrate a slight decrease in recognition accuracy on the MNIST, CIFAR-10, and CIFAR-100 datasets.

\section{Bipolar Morphological Networks}

Bipolar morphological (BM) neuron presented in \cite{limonova2019} performs multiplication-free approximation of a classical neuron to increase its computational efficiency for specialized devices.

\subsection{BM neuron model}

A classical neuron performs the following operation:
\begin{equation}
\label{eq:classical}
y(\mathbf{x}, \mathbf{w}) = \sigma \left (\sum_{i=1}^N w_i x_i + w_{N+1} \right),
\end{equation}
where $\mathbf{x}$ is an input vector of length 
$N$, $\mathbf{w}$ is weight vector of length $N+1$ and $\sigma(\cdot)$ is a nonlinear activation function. 

Calculations in the bipolar morphological neuron can be expressed as:
\begin{equation}
\begin{aligned}
\label{eq_bm}
y_{BM}(\mathbf{x}, V, v) = \sigma \left( \exp{\max_{j=1}^N (\ln x_j^+ + V_{j}^+)} - \right . \\
\left. - \exp{\max_{j=1}^N (\ln x_j^+ + V_{j}^-)} - \exp{\max_{j=1}^N (\ln x_j^- + V_{j}^+)} + \right .\\
\left. + \exp{\max_{j=1}^N (\ln x_j^- + V_j^-)} + v \right),
\end{aligned}
\end{equation}

\begin{equation}
x_j^+ = \begin{cases}
       x_j, x_j \geq 0, \\
       0, x_j < 0,
       \end{cases} \\
\end{equation}

\begin{equation}
x_j^- = \begin{cases}
       -x_j, x_j < 0, \\
       0, x_j \geq 0,
       \end{cases} \\
\end{equation}
where $\mathbf{x}$ is an input vector of length $N$, $V^+$, $V^-$ are weight vectors of size $N$, $v$ is bias, $\sigma(\cdot)$ is a nonlinear activation function.
We define $\ln 0 = -\infty$ and replace it by a big enough negative value for actual computations.

Since the neuron processes positive and negative parts of input $\mathbf{x}$ in a quite similar manner, we can interpret it as two identical computational paths responsible for excitation and inhibition.

We consider $V_{j}^+$ and $V_{j}^-$ as separate weights and train them independently.

\subsection{Training}
\label{sec:training}

The training method for networks with bipolar morphological layers is shown in \cite{limonova2019}.
The problem with the straightforward construction of the BM network and training it using standard gradient methods is that there is only one non-zero gradient element due to max operation, and only one weight is updated at each iteration. Thus it may show poor quality. Some weights can never be updated and never fire after training, thus giving redundancy to the network.

Instead, we can use incremental layer-by-layer conversion from the standard layer to the BM layer.
The approach is in training standard network and modifying convolutional and fully-connected layers from the first to the last and training new partly BM network.
It can be summarized as:
\begin{enumerate}
    \item Train classical network using conventional gradient descent-based methods;
    \item For each convolutional and fully-connected layers: replace neurons of type (\ref{eq:classical}) with weights $w$ by the BM-neurons with weights $\{V^+, V^-, v\}$, where:
    
\begin{equation}
\label{eq_convert}
\begin{aligned}
V_j^+ &=
    \begin{cases}
    \ln w_j , \mbox{ if } w_j > 0,\\
    -\infty , \mbox{ otherwise},
    \end{cases} \\
V_j^- &=
    \begin{cases}
    \ln |w_j| , \mbox{ if } w_j < 0,\\
    -\infty , \mbox{ otherwise},
    \end{cases}\\
v &= w_{N+1}.
\end{aligned}
\end{equation}

    \item Perform additional training of the network after conversion of each layer using same method as in 1.
\end{enumerate}

\subsection{Computational complexity}
\label{sec:comp_complexity}

Since neurons in neural network models are organized into layers, we consider a BM layer.
The layer uses many addition and maximum operations instead of multiplications.
However, the exact number depends on the computation organization.
Although the length of the $x^+$ and $x^-$ is $N$ for each of them, together, they have $N$ non-zero terms precisely.
Only these terms will contribute to the result.
So, we can say that the number of log operations is only $N$ and does not consider those zero terms.

The standard convolutional layer with input $I_{L\times M \times C}$ and output $O_{L\times M \times F}$ does the following:
\begin{equation}
\begin{aligned}
O(l, m, f) = \sigma \left( \sum_{c = 1}^C \sum_{\Delta l = 0}^{K-1} \sum_{\Delta m = 0}^{K-1} I(l + \Delta l, m + \Delta m, c)  \cdot \right. \\
\biggl . \cdot w(\Delta l, \Delta m, c, f) + b(f) \biggr ),~ f = \overline{1, F}, l = \overline{1, L}, m = \overline{1, M}
\end{aligned}
\end{equation}
Here $F$ is the number of filters, $C$ is the number of input channels, $K \times K$ is the spatial dimensions of the filter, input image size is $L \times M \times C$, $w$ is a set of convolutional filters, $b$ is the bias. We suppose $I$ is padded properly for the result to be of the same size.

The standard fully-connected layer with input $I(p)$ and output $O_Q$ does:
\begin{equation}
\begin{aligned}
O(q) = \sigma \left( \sum_{p = 1}^P I(p)  \cdot w(p, q) + b(q) \right ),~ q = \overline{1, Q}
\end{aligned}
\end{equation}
Here $P$ is the number of inputs, $Q$ is the number of neurons in the layer, $w$ is a set of fully-connected weights, $b$ is set of biases.

The number of operations for the standard and BM convolutional layers is shown in Table \ref{table1}. In Table \ref{table_fc}, we show the number of operations for standard and BM fully-connected layers.

\begin{table}[ht]
    \caption{The number of operations in the convolutional (conv) layer of BM and standard models. $F$ is the number of filters, $C$ is the number of input channels, $K \times K$ is the spatial dimensions of the filter, input image size is $L \times M \times C$.}
    \label{table1}
    \begin{small}
    \begin{center}
        \begin{tabular}{p{0.5cm}p{2.5cm}p{2.5cm}}
        \hline
        \hline
        Op & Standard conv & BM conv\\
        \hline
        \hline
        $\sigma(\cdot)$ & $F L M$ & $FL M$\\
        \hline
        Exp
        & 0 & $4 F L M$\\
        \hline
        Log & 0 & $C L M$ \\
        \hline
        Add & $F K^2 C LM$ & $2F(K^2 C+2)LM$\\
        \hline
        Max & 0 & $2 F(K^2 C-1)LM$ \\
        \hline
        Mul & $F K^2 C LM$ & 0\\
        \hline
        \hline
        \end{tabular}
    \end{center}
    \end{small}
\end{table}

\begin{table}[ht]
    \caption{The number of operations in the fully-connected (fc) layer of BM and standard models. $P$ is the number of inputs, $Q$ is the number of neurons in the layer.}
    \label{table_fc}
    \begin{small}
    \begin{center}
        \begin{tabular}{p{0.5cm}p{2.5cm}p{2.5cm}}
        \hline
        \hline
        Op & Standard fc & BM fc\\
        \hline
        \hline
        $\sigma(\cdot)$ & $Q$ & $Q$\\
        \hline
        Exp
        & 0 & $4Q$\\
        \hline
        Log & 0 & $P$ \\
        \hline
        Add & $QP$ & $2Q(P + 2)$\\
        \hline
        Max & 0 & $2Q(P - 1)$ \\
        \hline
        Mul & $QP$ & 0\\
        \hline
        \hline
        \end{tabular}
    \end{center}
    \end{small}
\end{table}

\subsection{Hardware implementation complexity}

In general, in order to compare the computational efficiency of such structures, we need to understand what kind of computing device is involved and know the characteristic latency and throughput of multiplier and adder.
On modern x86-64 and ARM architectures, for example, general-purpose Arithmetic Logic Units (ALUs) are used.
Execution time for the multiplication does not differ from the time for addition for floating-point data and only slightly differs for integer vector data (see Table \ref{table2}). 
Therefore, it will be extremely difficult to obtain inference acceleration implementing a BM network on a CPU, even with the coefficients and input signals converted to integers, since the total number of operations is more than in the standard layer.
For this reason, the proposed model is primarily aimed at FPGA and ASIC projects.

\begin{table}[ht]
    \caption{The latency and average throughput of vector arithmetic operations for 32-bit packed values in a vector \cite{AgnerFog, arm}.}
    \label{table2}
    \begin{small}
    \begin{center}
        \begin{tabular}{p{1.5cm}p{2.5cm}p{1.5cm}p{1.5cm}}
        \hline
        \hline
        Op &latency & throughput\\
        \hline
        \hline
        \multicolumn{3}{c}{Intel Skylake-X, floating-point 128-bit vector}\\
        \hline
        add & 4 & 0.5\\
        \hline
        max & 4 & 0.5-1\\
        \hline
        mul & 4 & 0.5-1\\
        \hline
        \multicolumn{3}{c}{Intel Skylake-X, integer 128-bit vector}\\
        \hline
        add & 1 & 0.33\\
        \hline
        max & 1 & 0.5\\
        \hline
        mul & 5 & 0.5\\
        \hline
        mul+add & 5 & 0.5\\
        \hline
        \multicolumn{3}{c}{ARM Cortex-A57, floating-point 128-bit vector}\\
        \hline
        add & 5 & 2\\
        \hline
        max & 5 & 2\\
        \hline
        mul & 5 & 2\\
        
        \multicolumn{3}{c}{ARM Cortex-A57, integer 128-bit vector}\\
        \hline
        add & 3 & 2\\
        \hline
        max & 3 & 2\\
        \hline
        mul & 5 & 1\\
        \hline
        mul+add & 5 & 1\\

        \hline
        \hline
        \end{tabular}
    \end{center}
    \end{small}
\end{table}

In this case, it is possible to make an efficient implementation with parallel execution of 4 computational paths with a specialized adder and maximum blocks, and not the general-purpose ALU. An inevitable delay will be caused by the multiplexing mechanism, which is necessary to direct the input signal to the desired computational pathway. 

Let us estimate the number of logic gates and clock cycle latency required for the arithmetical operations involved in the computations.
We have used Verilog HDL to get register transfer level description of addition, multiplication and maximum computation arithmetic units conforming to IEEE 754 floating-point standard, and Synopsys Design Compiler with 65 nm technologic libraries to implement it at gate-level and obtain logic gate complexity and latency characteristics.
For the logarithm and exponent operation, we have used software approximation through addition and multiplication to evaluate the hardware complexity. These values for single-precision data type are shown in Table \ref{table_gates}. 

Our custom implementation of the $\log$ function gives the precision of 4 decimal digits but is faster than the full-precision one. Let us describe it.

The floating-point numbers of single-precision in IEEE 754 are represented as the sign $s$, mantissa $b = b_0 b_1 b_2, \dots, b_{22}$ and exponent $e$:
\begin{equation}
x = 2^{e - 127} \cdot 1.b_{22} b_{21}, \dots, b_{0}
\end{equation}
So, 
\begin{equation}
\log_2 x = e - 127 + \log_2 (1 + b_{22} b_{21}, \dots, b_{0}),
\end{equation}
where $b_{22} b_{21}, \dots, b_{0}$ is in $[0, 1)$.
It means that we only need to approximate $\log_2 (1 + y)$ for $y \in [0, 1)$.

The approximation we construct is polynomial and has the 5$^{th}$ order:
\begin{equation}
f(y) = \log_2 (1 + y) \rightarrow \tilde{f}(y) = \sum_{i=0}^5 C_i y^i
\end{equation}

We obtain the coefficients $C_i$ by solving the system of linear equations. For 3 points 0, 0.5 and 1 we equate the values of $f(y)$ to the values $\tilde{f}(y)$. The same we do with the values of $f'(y)$ and $\tilde{f}'(y)$. The resulting $C_i$ are \{0, 1.44269504,  $-0.71249131$, 0.42046732,  $-0.1955884$, 0.04491735\}, and the approximation has a maximum error of about $7 \cdot 10^{-5}$ in $[0, 1)$. 
When computed with Horner's method, it uses only 5 multiplications and 6 additions (including the one to get $e - 127$) and bit manipulations to get $s, e$, and $b$, which are free for hardware. 

For the exponent we used a reference implementation for approximated exponent from Intel~\cite{intel}.

\begin{table}[ht]
    \caption{The estimate number of gates and latency for arithmetical operations}
    \label{table_gates}
    \begin{small}
    \begin{center}
        \begin{tabular}{p{1cm}p{2cm}p{3cm}}
        \hline
        \hline
        Op & Gates & Latency, clock cycles\\
        \hline
        \hline
        add & 16048 & 3\\
        \hline
        max & 1464 & 2\\
        \hline
        mul & 35345 & 4\\
        \hline
        log & 154179 & 35\\
        \hline
        exp & 256965 & 21\\
        \hline
        \hline
        \end{tabular}
    \end{center}
    \end{small}
\end{table}

\begin{figure}[!t]
\centering
\includegraphics[width=3in]{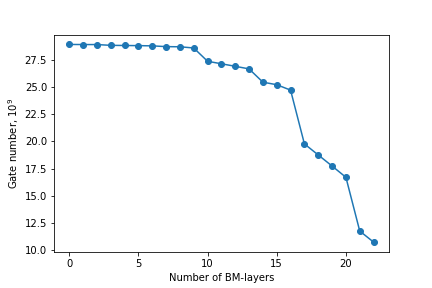}
\caption{The total gate number in convolutional layers of the ResNet-22 network depending on the number of BM convolutions. The rest of the network still uses standard convolutional layers.}
\label{fig:complexity}
\end{figure}

Knowing the number of single operations (see Section \ref{sec:comp_complexity}) and assuming that all four terms in (\ref{eq_bm}) are computed in parallel (twice less operation for add and max for one thread and 4 times less for exp) we obtain the approximate gate complexity of the circuit and its latency. The ratios of the gate numbers and the clock cycle latencies for the standard and BM convolutional layers are presented in Table \ref{table_estimate}. These ratios demonstrate that for core layers inside the network with quite a large number of input channels, we can get 2.1-2.9 times fewer gates and 15-30\% lower latency for the BM layer. Figure \ref{fig:complexity} illustrates the total gate number required for all convolutional layers for the ResNet-22 for CIFAR-10 (see Section \ref{sec:bm_resnet}) depending on how many standard layers are replaced by BM ones. Of course, the real ASICs do not require a separate unit for each layer, and we only demonstrate the general complexity decrease of the BM network.

\begin{table}[ht]
    \caption{The approximate gate number and latency ratios for standard and BM convolutional layers.}
    \label{table_estimate}
    \begin{small}
    \begin{center}
        \begin{tabular}{p{0.5cm}p{0.5cm}p{0.5cm}p{2cm}p{2cm}}
        \hline
        \hline
        $F$ & $C$ & $K$ & Gates standard/BM & Latency standard/BM\\
        \hline
        \hline
        16 & 1 & 1 & 0.16 & 0.22\\
        16 & 16 & 1 & 1.14 & 0.80\\
        32 & 1 & 1 & 0.17 & 0.23\\
        32 & 32 & 1 & 1.64 & 1.02\\
        64 & 1 & 1 & 0.17 & 0.23\\
        64 & 64 & 1 & 2.11 & 1.18\\
        128 & 1 & 1 & 0.17 & 0,23\\
        128 & 128 & 1 & 2.45 & 1.28\\
        256 & 1 & 1 & 0.17 & 0.23\\
        256 & 256 & 1 & 2.67 & 1,34\\
        512 & 1 & 1 & 0.17 & 0,23\\
        512 & 512 & 1 & 2.80 & 1.37\\
        
        16 & 1 & 3 & 1.02 & 0.87\\
        16 & 16 & 3 & 2.50 & 1.29\\
        32 & 1 & 3 & 1.03 & 0.89\\
        32 & 32 & 3 & 2.70 & 1.34\\
        64 & 1 & 3 & 1.03 & 0.89\\
        64 & 64 & 3 & 2.81 & 1.37\\
        128 & 1 & 3 & 1.04 & 0.91\\
        128 & 128 & 3 & 2.87 & 1.39\\
        256 & 1 & 3 & 1.04 & 0.90\\
        256 & 256 & 3 & 2.9 & 1.39\\
        512 & 1 & 3 & 1.04 & 0.90\\
        512 & 512 & 3 & 2.92 & 1.40\\
        \hline
        \hline
        \end{tabular}
    \end{center}
    \end{small}
\end{table}

However, there are several things that we can improve in the BM layer model for practical purposes. For example, we can use less precise logarithm approximation. Even more promising is using quantized BM layers to perform more calculations in integers. 

\section{BM-ResNet}

The ResNet \cite{he2016deep, he2016identity} is a modern deep neural network architecture.
The core idea is to use an identity shortcut connection to skip some layers to deal with vanishing gradients' problem.
It allows us to stack convolutional layers and obtain better recognition quality than shallower models.
So, ResNet is a scalable and accurate model that founds wide application in practice.

\begin{figure}[!t]
\centering
\includegraphics[height=6in]{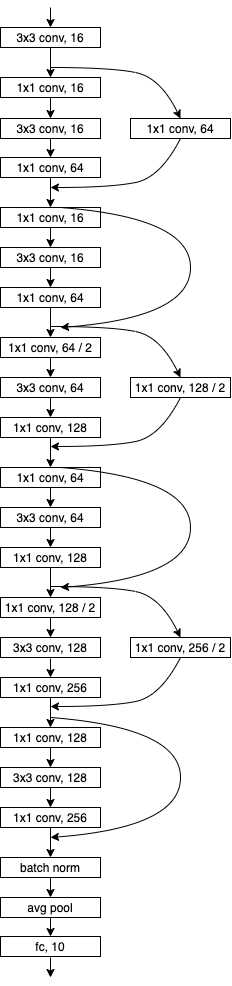}
\caption{The ResNet architecture with 22 convolutional layers used in experiments. Batch normalization and activations are omitted for simplicity.}
\label{fig:resnet}
\end{figure}

Our goal was to replace all classical neurons in convolutional layers by BM-neurons, keeping the network structure the same.
We conducted training according to the approach from Section~\ref{sec:training}.
We trained 22-layer ResNet-v2 (the architecture is briefly shown in Fig. \ref{fig:resnet}) at first with standard convolutional layers (step 1). 
Then we converted the convolutions layer by layer to BM convolutions according to (\ref{eq_convert}) (step 2) and trained the network for the 50 epochs (step 3).
The layers were converted sequentially from the first to the last.
After the conversion of all layers, the whole network was trained until the accuracy stopped improving.
The experiment is aimed at optimizing training time because the modelling time of BM-neuron using standard frameworks is quite slow.
We performed training with a standard Adam optimizer \cite{kingma2014adam} minimizing cross-entropy loss.

\subsection{MNIST}

\begin{figure}[!t]
\centering
\includegraphics[width=3in]{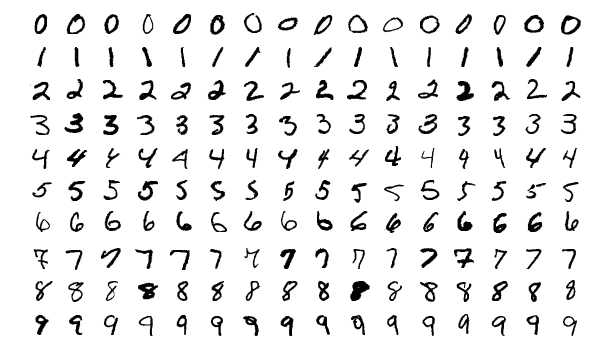}
\caption{Sample images from MNIST dataset.}
\label{fig:mnist}
\end{figure}

MNIST is a database of gray handwritten digits consisting of 60000 gray images of size $28 \times 28$ pixels for train and 10000 images for test~\cite{MNIST}.
We used 10\% of the training set for validation and the rest for training.
A few samples from the dataset are shown in Fig.~\ref{fig:mnist}.

\begin{figure}[!t]
\centering
\includegraphics[width=2.7in]{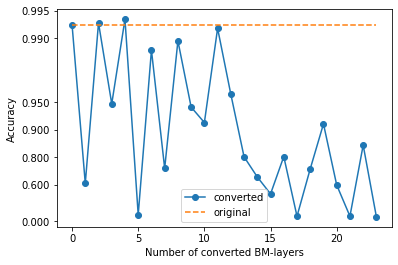}
\caption{Accuracy on MNIST after conversion before additional training.}
\label{fig:plot_mnist}
\end{figure}

\begin{figure}[!t]
\begin{minipage}[h]{\linewidth}
\centering
a) \includegraphics[width=2.7in]{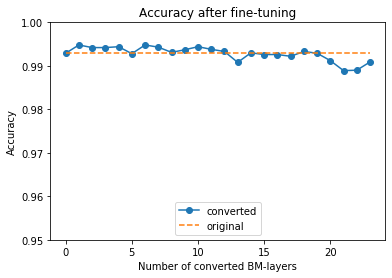}
\end{minipage}
\vfill
\begin{minipage}[h]{\linewidth}
b) \centering
\includegraphics[width=2.7in]{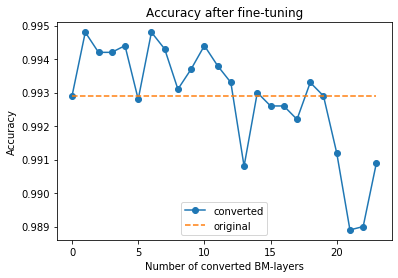}
\end{minipage} \centering
\caption{Accuracy on MNIST after conversion and additional training for accuracy range 0.95-1.00 (a) and 0.989-0.995 (b).}
\label{fig:plot_mnist_ft}
\end{figure}

\begin{figure}[!t]
\centering
\includegraphics[width=2.7in]{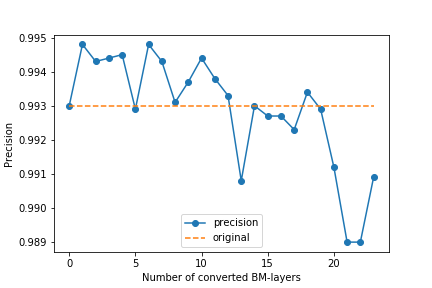}
\caption{Macro-average precision on MNIST after conversion and additional training.}
\label{fig:mnist_precision}
\end{figure}

\begin{figure}[!t]
\centering
\includegraphics[width=2.7in]{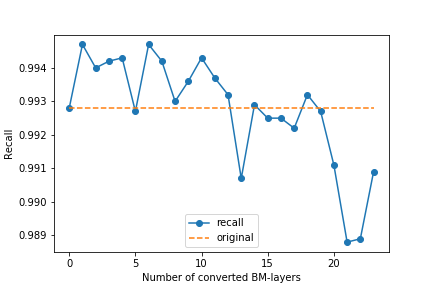}
\caption{Macro-average recall on MNIST after conversion and additional training.}
\label{fig:mnist_recall}
\end{figure}

The accuracy of the original ResNet was 99.3\%.
Evolution of the accuracies in the process of layer conversion to BM is shown in Fig.~\ref{fig:plot_mnist}-\ref{fig:plot_mnist_ft}. The final macro-average precision and recall are shown in Fig.~\ref{fig:mnist_precision} and Fig.~\ref{fig:mnist_recall}.
We can see that, in general, the network was able to preserve the original accuracy. 
Some fluctuations can be associated with not enough training time for each conversion step for the network to converge to the best accuracy. 
However, these layers were affected during further training, so it should not affect the final result with all BM convolutions.

The BM-ResNet demonstrated only slight accuracy decrease to 99.1\% (from 99.3\%) and can still be considered suitable for practical usage.

\subsection{CIFAR-10}
\label{sec:bm_resnet}

\begin{figure}[!t]
\centering
\includegraphics[width=2.5in]{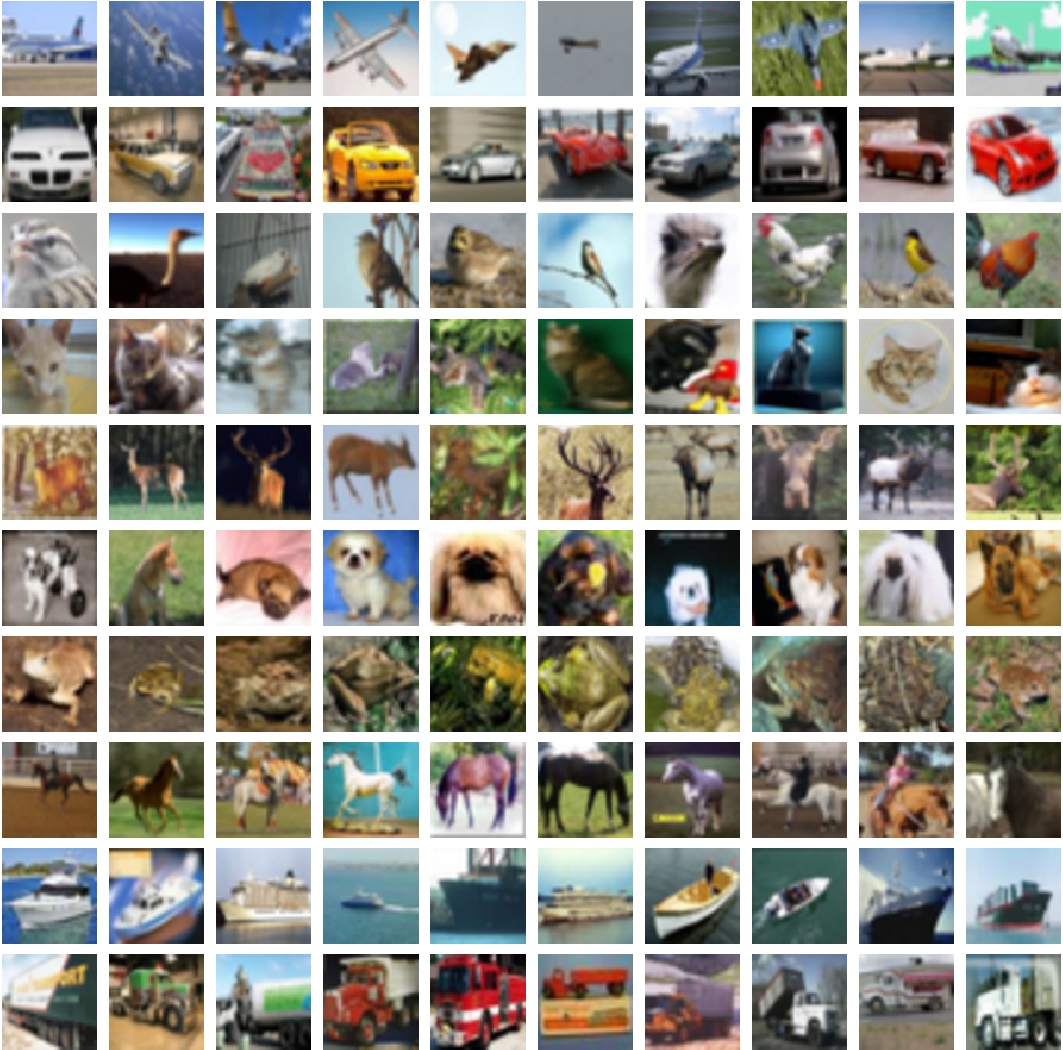}
\caption{Sample images from CIFAR-10 dataset.}
\label{fig:cifar10}
\end{figure}

CIFAR-10 is a database with 60000  $32 \times 32$ color images~\cite{CIFAR10} for train and 10000 images for test.
These images show objects of 10 different classes.
A few samples from the dataset are shown in Fig~\ref{fig:cifar10}.
We used standard preprocessing (normalized the pixel values of each sample to be in a range $[0, 1]$, and subtracted a mean image over the whole training database from each sample). 
We also used data augmentation, which included random horizontal and vertical shifts and random horizontal flips.

\begin{figure}[!t]
\centering
\includegraphics[width=2.5in]{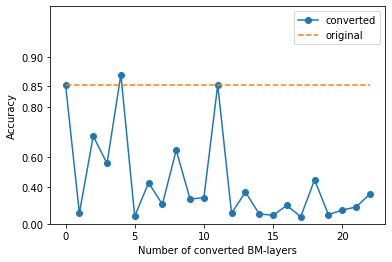}
\caption{Accuracy on CIFAR-10 after conversion before additional training.}
\label{fig:plot_cifar}
\end{figure}

\begin{figure}[!t]
\centering
\includegraphics[width=2.5in]{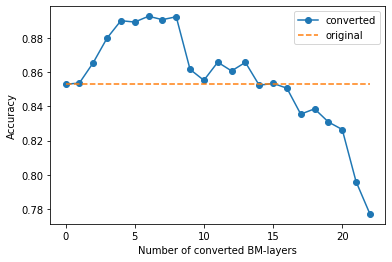}
\caption{Accuracy on CIFAR-10 after conversion and additional training.}
\label{fig:plot_cifar_ft}
\end{figure}

\begin{figure}[!t]
\centering
\includegraphics[width=2.8in]{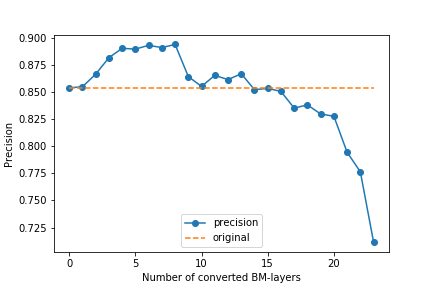}
\caption{Macro-average precision on CIFAR-10 after conversion and additional training.}
\label{fig:cifar10_precision}
\end{figure}

\begin{figure}[!t]
\centering
\includegraphics[width=2.8in]{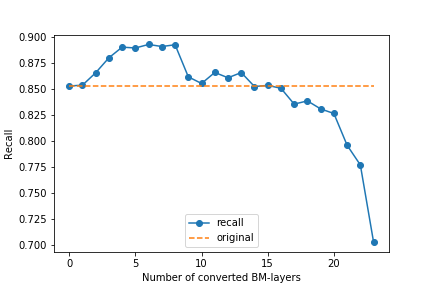}
\caption{Macro-average recall on CIFAR-10 after conversion and additional training.}
\label{fig:cifar10_recall}
\end{figure}

Evolution of the accuracies in the process of layer conversion to BM is shown in Fig.~\ref{fig:plot_cifar}-\ref{fig:plot_cifar_ft}. The final macro-average precision and recall are shown in Fig.~\ref{fig:cifar10_precision} and Fig.~\ref{fig:cifar10_recall}.
The accuracy of the standard ResNet was 85.3\%. The BM-ResNet with all the convolutional layers converted demonstrated accuracy decrease to 77.7\%, which is a significant accuracy drop. However, with 16 BM-convolutional layers, its accuracy was 85.1\%, and with 18 BM-convolutional layers, it was 83.9\%. 
So, the BM-ResNet with 16 converted layers is nearly as good as the original ResNet.
%We believe that the accuracy decrease of the fully transformed network is mainly caused by our training approach's imperfection, not the lack of expressive power of the BM-network.

\section{Conclusion}

In this paper, we managed to obtain a 22-convolutional layer ResNet-like with bipolar morphological convolutions.
We converted and trained the network to classify MNIST and CIFAR-10 datasets.
The accuracy on MNIST was 99.3\% with standard convolutional layers and 99.1\% with morphological ones. The accuracy on CIFAR-10 was 85.3\% with standard convolutional layers and 85.1\% with 16 BM-convolutional ones.

Since FPGA implementation of our BM networks requires about 2.1-2.9 times fewer gates and gets 15-30\% lower latency for large enough layers, our results show that it is possible to create a neural processing unit with lower power consumption, higher inference speed and reasonable accuracy compared to units for standard networks.

Our experimental setup for training can be found at: \url{https://github.com/SmartEngines/bipolar-morphological-resnet}

\section*{Acknowledgment}
The authors would like to thank A.B. Merkov, PhD, for his helpful advice and comments. 
This work is partially supported by Russian Foundation for Basic Research (projects 17-29-03240, 18-07-01384).

%listoftodos    %%Merx

%\IEEEtriggeratref{25}

\bibliographystyle{IEEEtran}
\bibliography{biblio}

\end{document}